\documentclass[letterpaper]{article} 
\usepackage{aaai2026}  
\usepackage{times}  
\usepackage{helvet}  
\usepackage{courier}  
\usepackage[hyphens]{url}  
\usepackage{graphicx} 
\usepackage{natbib}  
\usepackage{caption} 
\usepackage{mathtools}
\usepackage{algorithm}
\usepackage{algorithmic}
\nocopyright
\usepackage{newfloat}
\usepackage{listings}
\DeclareCaptionStyle{ruled}{labelfont=normalfont,labelsep=colon,strut=off} 
\floatstyle{ruled}
\newfloat{listing}{tb}{lst}{}
\floatname{listing}{Listing}
\title{Novelty-based Tree-of-Thought Search for LLM Reasoning and Planning}
\author{
    Leon Hamm\textsuperscript{\rm 1},
    Zlatan Ajanovic\textsuperscript{\rm 1}\thanks{Work done while at RWTH Aachen}
}
\affiliations{
    \textsuperscript{\rm 1}RWTH Aachen\\
    Aachen, Germany
}

\usepackage{bibentry}

\usepackage{amsmath}
\usepackage{cleveref}
\usepackage{tabularx}
\usepackage{amssymb}
\usepackage{booktabs}
\usepackage{multirow}

\newcolumntype{C}{>{\centering\arraybackslash}X}

\begin{document}

\maketitle

\begin{abstract}
Although advances such as chain-of-thought, tree-of-thought or reinforcement learning have improved the performance of LLMs in reasoning and planning tasks, they are still brittle and have not achieved human-level performance in many domains, and often suffer from high time and token costs.
Inspired by the success of width-based search in planning, we explore how the concept of novelty can be transferred to language domains and how it can improve tree-of-thought reasoning. A tree of thoughts relies on building possible “paths” of consecutive ideas or thoughts. These are generated by repeatedly prompting an LLM. In our paper, a measurable concept of novelty is proposed that describes the uniqueness of a new node (thought) in comparison to nodes previously seen in the search tree. Novelty is estimated by prompting an LLM and making use of embedded general knowledge from pre-training. This metric can then be used to prune branches and reduce the scope of the search. Although this method introduces more prompts per state, the overall token cost can be reduced by pruning and reducing the overall tree size. This procedure is tested and compared using several benchmarks in language-based planning and general reasoning.
\end{abstract}


\section{Introduction}

Large language models (LLMs) have attracted considerable attention for their 
performance across a wide range of language-based tasks, including exams and 
coding~\cite{GPT4TechReport, LiveBench}, and even display limited reasoning 
capabilities~\cite{EvalReasoningChatGPT, GLoRE, CoTReasoning}. However, these 
models struggle with complex reasoning and long-term planning when used 
naively~\cite{PlanningLLMCriticalInvestig, CausalParrots}. More structured 
approaches such as Tree of Thoughts (ToT)~\cite{ToT} partially address this, 
but suffer from exponential runtime and token costs due to branching 
search~\cite{ToS_Effic}. Training-based alternatives such as reinforcement 
learning from human feedback require expensive fine-tuning~\cite{rlhf-llm, 
deepseek-R1}, and overly long chains of thought can even degrade 
performance~\cite{more_is_less_cot}.

Classical planning algorithms, by contrast, have long handled such problems 
efficiently. Iterative Width (IW)~\cite{OrigWidth, WidthPlanning} achieves 
strong performance by pruning states according to a novelty measure, with 
runtime exponential only in the problem's width. Since many problem domains 
have been shown to have low width, this is highly effective in practice. We 
hypothesize that the same holds for language-based planning tasks, but 
defining an analogous notion of novelty in the language domain remains an 
open challenge. Classical planners also typically require problems to be 
formalized in a rigid structure, limiting their applicability to natural 
language settings.

In this work, we apply novelty-based pruning within a ToT framework for 
planning and reasoning tasks. We define a planning algorithm comprising four 
sub-tasks: action generation, successor state mapping, plan verification, and 
novelty estimation. Each sub-task and the overall algorithm are evaluated 
using a setup inspired by the PlanBench~\cite{PlanBench} benchmark, enabling 
direct comparison with classical planners. All code, prompts, and results are 
available in the accompanying GitHub repository\footnotemark.
\footnotetext{github.com/anonymous/redacted}

\section{Background}
\subsection{Classical Planning}

Classical planning is an important framework for reasoning about actions~\cite{PracticalPlanningBook}, 
originating in the 1970s with the introduction of STRIPS~\cite{STRIPS}. The 
field has been continuously researched since, with a notable milestone being 
the composition of the Planning Domain Definition Language 
(PDDL)~\cite{PDDL} in 1998, a community standard whose variants remain in 
use to this day.

The general approach assumes a state-based representation of the world, where 
states are characterized by a collection of atoms. We 
adopt the notation and problem schema of \citeauthor{OrigWidth}. The set of 
all possible states is denoted $S$, with an initial state $s_0 \in S$ serving 
as the starting point. An agent may choose from a set of actions $O$ that 
transition the current state $s \in S$, subject to constraints $A(s) \subseteq 
O$ that restrict which actions are admissible in a given state. Any action 
$a \in A(s)$ transforms the current state according to a successor function 
$s' = f(a, s)$. The goal $S_G \subseteq S$ is defined as the set of states 
satisfying certain target conditions. The planning problem then consists of 
finding a sequence of admissible actions $a_1, \ldots, a_m$, called a plan, 
that maps the initial state to a goal state: the actions induce a series of 
states $s_0, s_1, \ldots, s_m$ where $a_i \in A(s_i)$, $s_{i+1} = 
f(a_i, s_i)$, and $s_m \in S_G$. Such a plan may additionally be optimal 
with respect to criteria such as length. The planning problem can thus be 
expressed as a tuple $\langle S, s_0, S_G, A, f \rangle$.

Most classical planners, such as Fast Forward~\cite{FFplanner} and Fast 
Downward~\cite{FastDownward}, employ heuristic search to find such optimal 
plans.

\subsection{Width-Based Planning}
Lipovetzky and Geffner introduced a novel framework for characterizing the 
complexity and hardness of planning problems~\cite{OrigWidth, WidthPlanning}. 
Their approach defines a notion of \textit{width} that bounds the complexity 
of a problem, built on the concept of state novelty. Each state $s$ is 
represented as a subset of boolean features $\Phi(s) \subseteq F$, where $F$ 
denotes the set of all features defined by the problem. Planning is then 
modeled as a tree search over the graph induced by the initial state $s_0$ 
and its possible successors. The novelty $w(s)$ of a state is defined as the 
size of the smallest subset of features that are true in $s$ but not 
collectively true in any previously visited state $S^-_s$:

\begin{gather}
    w(s) = \min_{\substack{\phi \subseteq \Phi(s),\, \phi \nsubseteq 
    \Phi(s'), s' \in S^-_s}} |\phi|\text{.}
\end{gather}

Width-based planners exploit this notion by pruning any state whose novelty 
exceeds a specified threshold $k$. The width $w(P)$ of a planning problem is 
then defined as the minimum novelty required to solve it via breadth-first 
search under this pruning scheme. A width-based planner such as IW can find 
an optimal plan for a problem of width $k$ in $O(|F|^k)$. Notably, most 
problems with a single goal fluent have width at most two, yielding quadratic 
or even linear runtimes. Combined with problem serialization, this provides an 
efficient planner capable of quickly and optimally solving problems across a 
wide range of domains.

\subsection{Large Language Models}
Natural language processing has been an active research area since the 1950s 
and 60s~\cite{NLP}, but has advanced substantially with the adoption of deep 
learning and neural networks over classical statistical modeling. The 
introduction of pre-trained models based on the scalable transformer 
architecture~\cite{Attention} has greatly accelerated progress, with 
sufficiently large models exhibiting emergent abilities not seen in smaller 
counterparts~\cite{EmergentLLM}. These models, now the subject of extensive 
research~\cite{LLMSurvey}, are known as large language models (LLMs). Beyond 
excelling at tasks such as conversation and exams, LLMs can learn in-context, 
adapting to new problems using information provided at test time rather than 
relying solely on training data. Research has further shown that how a model 
is prompted can severely impact performance~\cite{EvalLLMSurvey, CoTReasoning}, 
giving rise to the field of prompt engineering.

We adopt a notation close to that of Yao et al.~(2023). A pre-trained language 
model $p$ with frozen parameters receives a sequence of prompt tokens 
$x = x[1], \ldots, x[n]$ and autoregressively predicts subsequent tokens by 
sampling from a conditional distribution: $x[n+1] \sim p(x[n+1]|x)$. 
Repeating this process yields a completion 
$y = y[1], \ldots, y[m] = x[n+1], \ldots, x[n+m]$, written concisely as 
$y \sim p(y|x)$, where $x$ and $y$ are natural language texts encoded as 
tokens. When prompting techniques are applied, inputs are first transformed 
via a named function $t^{\text{name}}$ before completion: 
$y \sim p^{\text{name}}(y|x) = p(y|t^{\text{name}}(x))$. In the simplest 
setting, the problem serves as the prompt and the completion as the solution. 
This can be improved with few-shot prompting, where example problem-solution 
pairs are prepended to the prompt~\cite{gpt3}, though such techniques have 
been criticized for brittleness and sensitivity to prompt 
design~\cite{BrittlePromptingLLM}, even if autonomous prompt generation is 
also possible~\cite{AutoFewShotPrompting}.
To avoid over-reliance on domain-specific prompts, we avoid few-shot prompting wherever possible.

\subsection{LLMs for Reasoning and Planning}
Despite being designed for text generation, LLMs can solve some reasoning 
tasks through autoregressive prediction alone. However, they tend to produce 
answers without deliberation, analogous to a human responding with the first 
thing that comes to mind, leading to poor performance on tasks requiring 
complex reasoning or long-term planning~\cite{LLMcantPlan}.

This has prompted researchers to investigate their use in planning 
tasks~\cite{PlanningSurveyLLM, LMzeroshotPlanners, DoAsICanNotAsISay} and to 
develop methods that introduce more deliberate reasoning. Two main approaches 
to LLMs in planning have emerged. The first combines LLMs with conventional 
symbolic planners or verifiers, using the model for tasks such as translating 
planning goals between natural language and structured planning 
languages~\cite{NatLang2PlanGoalLLM, LLM+P} or as a decision-making 
heuristic~\cite{LLMAssist, LLMplanner, DynamicPlanningLLM}. The second omits 
a conventional planner entirely, relying on the model and corrective measures 
alone~\cite{LMzeroshotPlanners, CorRePromptingPlanning}. While both approaches 
improve upon a naive LLM baseline, the latter has been criticized for 
over-reliance on in-context examples and poor generalization~\cite{LLMcantPlan, 
CoThoughtlessnessPlanning}.

To address the lack of deliberation more generally, methods that scale 
test-time compute through structured reasoning have been developed. Chain of 
Thought (CoT)~\cite{CoTReasoning} inserts explicit reasoning steps between 
example inputs and outputs, significantly boosting performance despite its 
simplicity. Its main limitations are the linearity of the reasoning process 
and the need for human-designed examples, which hinders domain transfer. Tree 
of Thoughts (ToT) and Graph of Thoughts~\cite{GoT} address the linearity 
issue by structuring reasoning as a tree or graph, with the LLM also serving 
as an evaluation heuristic for search. While effective, their branching factor 
leads to high compute costs~\cite{ToS_Effic}. More recently, large reasoning 
models such as OpenAI's o1~\cite{o1systemcard} are fine-tuned via 
reinforcement learning to reason implicitly step by step, though at 
substantial compute cost.

\subsection{Tree of Thoughts for Planning}
We employ ToT~\cite{ToT} as the planning problem solver, interpreting each 
thought as a planning state $s$, with the initial state $s_0$ as the root of 
the search tree. Any state is represented as natural language text used to 
prompt the model, and the goal is to find a sequence of thoughts via tree 
search that constitutes a valid solution. In this framework, the LLM is 
responsible for four sub-tasks: action generation, successor state mapping, 
plan verification, and novelty estimation. The first three form the baseline 
ToT and are described below; novelty estimation is the main novel contribution 
and is discussed in the following section. Note that explicit action generation 
and successor state mapping are not part of the original ToT formulation, but 
are introduced here to enable novelty estimation over states.

\subsubsection{Action Generation}
The LLM acts as an action generator, using its knowledge of the problem to 
propose possible next actions from a given state. We use sampling, prompting 
the model $m$ times independently:
\begin{gather}
    a_{i+1}^1 \sim p^{\text{sample}}(a_{i+1}|s_i),\ldots,
    a_{i+1}^m \sim p^{\text{sample}}(a_{i+1}|s_i)\text{.}
\end{gather}
\subsubsection{Successor State Mapping}
The LLM then maps each action to a successor state via a prompted successor 
function:
\begin{gather}
    s_{i+1} = f(a_{i+1}, s_i) \sim p^{\text{successor}}(s_{i+1}|a_{i+1}, s_i)\text{.}
\end{gather}
Various prompting strategies, such as CoT, can be applied here. The resulting 
states are stored for use in novelty estimation.

\subsubsection{Plan Verification}
At each step, the model verifies whether the current state satisfies the goal. 
This can be done either algorithmically, by parsing the state directly, or by 
prompting the LLM when no verifier is available. The former is domain-dependent 
and requires constraining the model's output format; the latter is more general 
but less reliable due to the black-box nature of LLMs.

\section{Novelty-based Pruning in Tree-of-Thoughts}
As established, many classical planning problems exhibit low width when serialized, suggesting inherent tractability. Yet despite this, LLMs continue to struggle with such problems~\cite{PlanningLLMCriticalInvestig, OrigWidth}. We aim to address this gap by building on an existing method (Tree of Thoughts) and explicitly exploiting low planning width to improve performance.

ToT works by constructing and searching a tree using a chosen search algorithm to find a solution to the given problem, an approach that is, in principle, applicable to almost any reasoning or planning task. Its key disadvantage, however, is the substantial time and token cost associated with LLM-driven tree search. The central question this work investigates is whether and how this search can be improved by pruning states according to a measure of novelty. 

To prune states efficiently, the concept of novelty must be adapted to work 
with unstructured natural language input, removing the need to define explicit 
atoms for each state and eliminating any requirement for the LLM to generate 
completions in a prescribed format. This greatly simplifies the overall 
approach. At the same time, the adapted notion of novelty must remain 
sufficiently analogous to the original to enable worthwhile pruning, and must 
be domain-agnostic, functioning as intended for any domain with states $S$.
We assume that states in any given domain have latent features $\Phi(s)$. 
Given a set of previously visited states $S^-_s \subseteq S$ and a new state 
$s \in S$, the adapted novelty measure $n(s|S_s^-)$ should correlate 
positively with the original novelty.\\

The classical planning novelty-search approach quantifies state novelty as an 
integer value relative to the search tree. However, preliminary tests showed 
that LLMs struggle to map novelty differences to integer values, and in many 
cases a classical integer-based notion of width or novelty is difficult to 
define or apply. A more general approach is therefore needed.

Instead, the LLM is queried about whether a new state is novel as a binary 
yes/no question, effectively pruning the search tree directly according to the 
model's internalized notion of novelty. This offers considerable flexibility, 
as the model can leverage its semantic understanding to, for example, assign 
greater importance to more salient features. The simplicity of this approach 
is furthermore beneficial in terms of both reliability and cost. States that 
are too large to process directly can additionally be summarized in a 
preliminary prompt step.

In contrast to other approaches, ours is designed to be domain-independent 
wherever possible. Consequently, no few-shot examples are used; the only 
domain knowledge provided is a hand-crafted context prompt explaining the 
rules of the domain. The LLM is then employed as a novelty estimator, using 
a simple prompt template alongside a running list of previously visited states. 
Since all states are represented as strings, a prompt for boolean novelty 
estimation can be kept exceedingly simple:

\begin{listing}[h]
\caption{Prompt for Novelty estimation}%
\label{lst:prompt-novelty}%
Is '{new\_state}' a novel state compared to this list of states 
'{previous\_states\_str}'? Respond with 'yes' or 'no'. Do not add any 
other text.    
\end{listing}

Contextual details such as the query state and goal atoms can be 
programmatically appended to the instructions. To preserve generalizability, 
we deliberately avoid domain-specific prompt engineering for novelty 
estimation. Instead, performance is improved by testing a range of 
domain-agnostic prompt templates.

\subsection{Direct Successors vs. State-Action Mapping}
Two options are provided for building the search tree. The first follows the 
approach described above, querying separately for the next action and then 
mapping that action to a successor state. The second combines action generation 
and successor state mapping into a single step, making it more general since 
not all domains have sensible definitions for these as separate tasks. In both 
options, plan verification is generalized to accept any sufficient goal 
condition, not just a classical planning goal. Novelty estimation can further 
be performed by providing the model with previous abstract steps in place of 
explicit states.

\subsection{BFS vs. DFS}
The tree can be searched using either breadth-first search (BFS) or 
depth-first search (DFS). While IW traditionally employs BFS, large action 
spaces can lead to exponential cost growth, which is particularly prohibitive 
in the context of LLM-based search. This renders BFS especially impractical 
when using ToT without pruning. The trade-offs between both strategies are 
examined in the evaluation.

\section{Evaluation Framework}
As outlined previously, the LLM is responsible for four sub-tasks: action 
generation, successor state mapping, plan verification, and novelty estimation. 
Evaluating each sub-task independently provides insight into both the 
individual capabilities of LLMs in planning and the overall approach.

The evaluation framework is inspired by PlanBench~\cite{PlanBench} and uses a 
PDDL engine to generate an arbitrary number of test instances, with classical 
planning providing well-defined ground truths. Bidirectional translation 
between PDDL and natural language is handled via algorithmic parsing, and 
model responses are evaluated using a combination of this translation and LLM 
queries.

\subsubsection{Action Generation}
For each test instance, a random planning problem is generated and the 
simulator produces the set of all valid actions from the current state. The 
LLM is prompted to enumerate these actions, given the state either in PDDL 
form or natural language, and its output is compared against the ground truth.

A second variant queries the model for a single next action given the goal, 
testing its ability to identify good actions rather than merely valid ones. The 
action is then assessed for both validity (admissibility under Blocksworld 
rules) and optimality (whether it is the first action of an optimal plan).

\subsubsection{Successor State Mapping}
Using the states and actions from the previous sub-task, the LLM is prompted 
to map each action to its resulting state, which is then compared to the 
simulator's output. A variant also tests mapping actions to states 
individually rather than jointly in a single prompt.

\subsubsection{Plan Verification}
During a ToT run, the LLM must determine whether a given sequence of actions 
constitutes a complete and valid plan. Half of the test instances contain valid 
plans generated by FastDownward~\cite{FastDownward}; the other half contain 
invalid plans generated randomly, guaranteed not to reach a goal state. The 
model is prompted with an origin state and a plan and must decide whether all 
goal conditions are satisfied.

\subsubsection{Novelty Estimation}
For each test instance, a complete IW search is conducted to determine the 
instance width, defined as the minimum novelty threshold at which a goal state 
is reached. A partial width-based search is then run with classical novelty 
pruning, stopped at a random point to produce a realistic intermediate search 
state. The LLM is prompted to decide whether the current state is novel or 
should be pruned, with the ground truth taken from the last iteration of IW.

Since most states in this setting are pruned as duplicates rather than due to 
novelty, a harder variant is also included in which all query states are 
non-duplicate and would be pruned strictly on the basis of exceeding the 
instance width. This stress-tests the model's ability to reason about novelty 
in the more challenging and informative pruning scenario.

\section{Experiments and Results}
We evaluate three main aspects: the applicability and potential of our 
approach as measured by problem hardness, the performance of the devised 
sub-tasks relative to their classical counterparts, and the overall 
performance of our approach compared against a naive LLM baseline and a 
standard ToT baseline.

\subsection{Applicability and Potential}
To assess the applicability of our approach, we compute the hardness of a 
reasoning problem from the original ToT paper in terms of latent novelty. 
Since many common planning domains have been shown to have width below three 
when serialized, and many language-based reasoning problems can be modeled as 
classical planning problems, we hypothesize that many LLM reasoning problems have a low inherent width and that a low latent width and a high volume of pruneable states will lead to more effective pruning in our LLM-based approach.

\subsubsection{Game of 24}
We examine the Game of 24, in which the solver is given four numbers 
$x_1,x_2,x_3,x_4 \in \mathbb{N}$ and must combine them using the four basic 
arithmetic operations to reach 24, using each number and intermediate result 
exactly once. A state is defined as the multiset of remaining numbers $R_i$, 
with the initial state $R_0 = \{\!\{x_1,x_2,x_3,x_4\}\!\}$. An action 
combines two numbers via an arithmetic operation, and the successor state 
replaces them with the result. To simplify goal specification, we augment the 
state with the remaining count $r = |R_i|$, yielding a goal expressible with 
two atoms. Formally:
\begin{gather*}
    S = \{\!\{\mathbb{Q}\}\!\}\footnotemark \times \mathbb{N} \\
    s_0 = (\{\!\{x_1,x_2,x_3,x_4\}\!\}, 4) \\
    s_i \eqqcolon (R_i,r) \\
    \Lambda = \{+,-,\cdot,\div\} \\
    O = \mathbb{Q} \times \mathbb{Q} \times \Lambda \\
    A(s_i) = A((R_i,r)) = \{(y_1,y_2,\lambda) \; | \; y_1 \in R_i,\\ y_2 \in R_i \setminus \{y_1\}, \: \lambda \in \Lambda, \: (\lambda = \div) \rightarrow (y_2 \neq 0), \: r > 1\} \\
    f(a,s_i) = f((y_1,y_2,\lambda),(R_i,r)) \\ = (\{y_1 \lambda y_2\} \uplus R_i \setminus \{y_1,y_2\}, r-1) \\
    S_G = \{\,(\{\!\{24\}\!\},1)\,\} \text{.}
\end{gather*}
\footnotetext{Multiset notation: $\{\!\{S\}\!\}$ denotes a multiset over 
underlying set $S$; $\uplus$ denotes multiset union.}

\begin{figure}[t]
  \centering
  \includegraphics[width=0.45\textwidth]{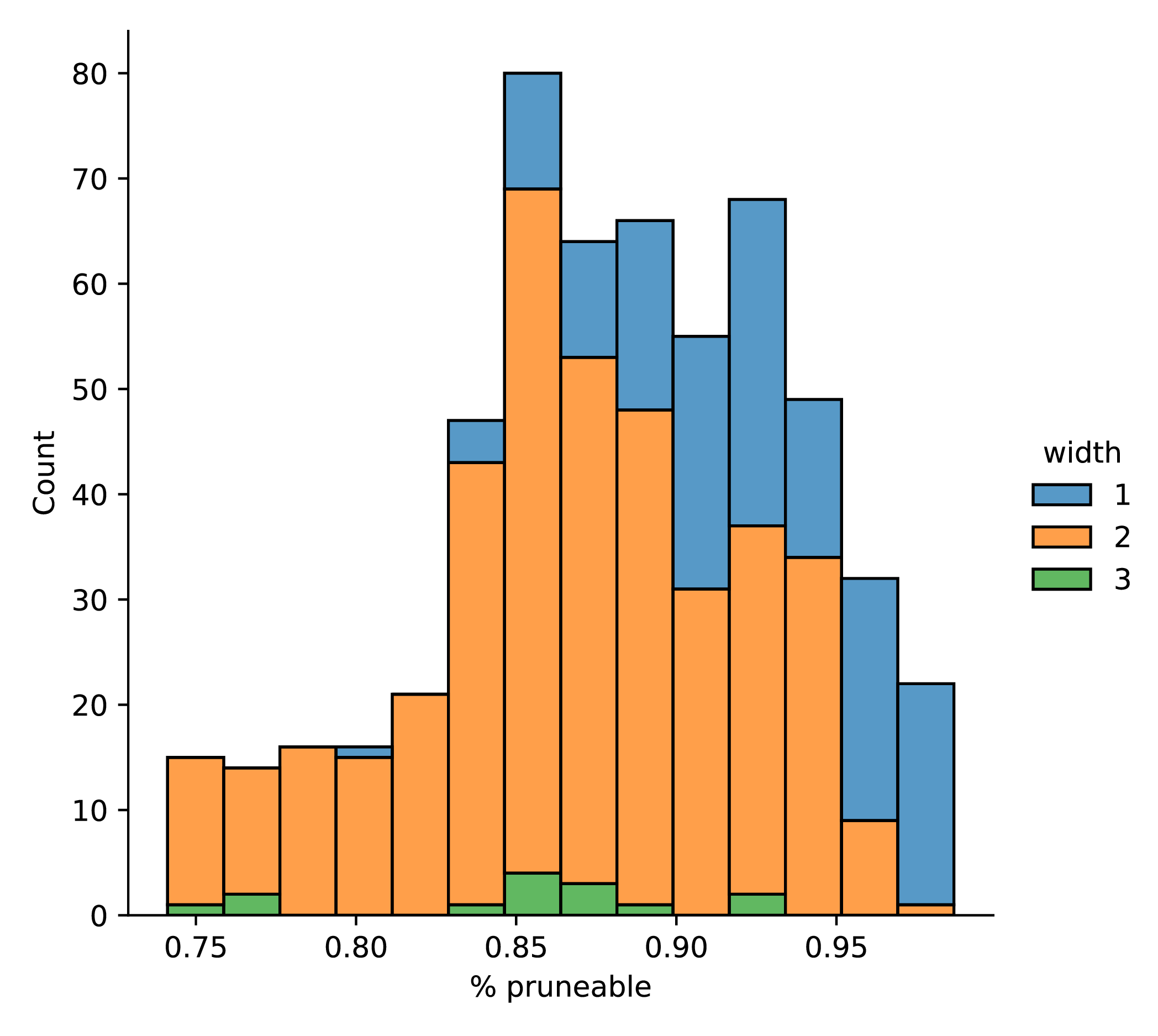}
  \caption{Distributions of computed widths and percentage of pruneable states 
  under width-based search for all examined Game of 24 instances.}
  \label{fig:widths-pruneables}
\end{figure}

The Game of 24 has an average effective width of $1.74$, with no instance 
exceeding width $3$ and $88.2\%$ of states pruneable on average. State trees 
contained an average of $3698$ states. The distributions are shown in 
Figure~\ref{fig:widths-pruneables}. These results confirm that a benchmark 
commonly used to test LLM reasoning shares the low-width structure of 
classical planning domains, establishing the theoretical basis for our 
approach. We next evaluate the LLM's ability to execute the required sub-tasks.

\subsection{Sub-task Evaluations}
The sub-tasks were evaluated using the Qwen3 14B model. As an open-source model small enough to be run locally, it is relatively fast while remaining competitive with other state-of-the-art models of a similar size, making it ideal for this testing.

All four sub-tasks were tested across two primary conditions:
\begin{itemize}
    \item \textbf{Prompting Style:} Standard PDDL-like versus Natural Language translation. For example, "(ontable b)" becomes "the blue block is on the table". Equivalent transformations are applied to actions to determine if the LLM performs better with structured, high-density information or sparse natural language.
    \item \textbf{Thinking Mode:} The model's built-in reasoning/thinking tokens were either toggled on or off.
\end{itemize}
The results can be found in \Cref{tab:subtask-results}.\\

\begin{table*}[!ht]
  \small
  \centering
  \begin{tabularx}{\textwidth}{l *{4}{C}}
    \toprule
    \multirow{2}{*}{\textbf{Sub-task}} 
      & \multicolumn{2}{c}{\textbf{Standard Prompting}} 
      & \multicolumn{2}{c}{\textbf{Natural Language Prompting}} \\
    \cmidrule(lr){2-3} \cmidrule(lr){4-5}
    & Normal & Thinking & Normal & Thinking \\
    \midrule
    Action Generation                & 1/50  & \textbf{50/50} & 1/50  & 48/50 \\
    Action Generation Single Valid   & 22/50 & 41/50          & 23/50 & \textbf{43/50} \\
    Action Generation Single Optimal & 12/50 & 33/50          & 12/50 & \textbf{34/50} \\
    Success State Mapping            & 15/50 & 38/50          & 34/50 & \textbf{44/50} \\
    Success State Mapping Separate   & 26/50 & \textbf{47/50} & 32/50 & 45/50 \\
    Goal State Verification          & 47/50 & 47/50          & \textbf{48/50} & \textbf{48/50} \\
    Novelty Pruning                  & 31/50 & \textbf{50/50} & 26/50 & \textbf{50/50} \\
    Novelty Pruning NDAP             & 0/50  & 0/50           & 0/50  & \textbf{1/50} \\
    \bottomrule
  \end{tabularx}
  \caption{Performance of LLM in Sub-tasks across prompting and thinking.}
  \label{tab:subtask-results}
\end{table*}

In most cases, thinking mode either improves upon or has similar results to its counterpart.
This matches expectations, as the model was specifically trained to process inputs better using thinking tokens.

Natural language prompting also mostly improves or keeps performance.
Standard PDDL prompting may perform better sometimes because it is more concise and easier to handle when working with a large number of states and atoms.

Interestingly, in thinking mode, the model performs better when asked to name all possible actions in Action Generation than when identifying the "next action to solve the problem" in Action Generation Single. The added goal of choosing the single correct action leads to decreased validity scores, though it provides more information regarding optimality.

Although the model correctly prunes in the Novelty Pruning sub-task, it completely fails at the more difficult variant.
This is due to good duplicate recognition but otherwise poor understanding of more complex novelty.

Other than that, the LLM is able to complete most sub-tasks successfully, especially when configured optimally.
The best results are usually achieved using natural language and thinking mode.

To uncover some of the reasons for lower performance scores, an iterative qualitative analysis is conducted.
It revealed a specific recurring failure point: the confusion between "pick-up" and "unstack" actions. According to the domain rules, blocks may only be picked up from the table; if stacked, they must be unstacked. The model struggles with this distinction, severely lowering Action Generation scores. 

To alleviate this, the context prompt was manually extended with a targeted example explaining this distinction (\Cref{lst:prompt-extension}).
\begin{listing}[tb]
\caption{Prompt Extension for Action Generation}%
\label{lst:prompt-extension}%
For example, if I have a blue block on the table and a red block on top of it (and clear), then I may NOT pick up the red block because it is on top of another block. This means I must instead unstack it.
\end{listing}
As shown in \Cref{tab:action_gen_prompt_diff}, this manual prompt extension yielded an apparent performance increase, showcasing the fragility of LLM prompting and the outsized impact of targeted examples.
In fact, even the prompts used for general sub-task evaluation prior to the extension were found by iterative testing of many different prompts.
Bad prompts often lead to a complete collapse of performance.

\begin{table*}[!ht]
  \small
  \centering
  \begin{tabularx}{\textwidth}{l *{4}{C}}
    \toprule
    \multirow{2}{*}{\textbf{Sub-task}} 
      & \multicolumn{2}{c}{\textbf{Standard Prompting}} 
      & \multicolumn{2}{c}{\textbf{Natural Language Prompting}} \\
    \cmidrule(lr){2-3} \cmidrule(lr){4-5}
    & Normal & Thinking & Normal & Thinking \\
    \midrule
    Action Generation Single Valid   & 39/50 & 45/50 & 40/50 & \textbf{50/50} \\
    Action Generation Single Optimal & 27/50 & 41/50 & 27/50 & \textbf{46/50} \\
    \bottomrule
  \end{tabularx}
  \caption{Performance of LLM in Action Generation Single with Prompt Extension.}
  \label{tab:action_gen_prompt_diff}
\end{table*}

\subsection{Complete Approach Evaluation}
Finally, the performance of the entire approach is measured on benchmarks in classical planning and LLM reasoning, comparing our method against ToT baselines across cost, accuracy, robustness, and adaptability. The primary objectives are to:
\begin{itemize}
    \item improve upon the token cost of regular ToT;
    \item retain high accuracy relative to state-of-the-art approaches;
    \item work towards the robustness of classical methods;
    \item make use of LLM flexibility.
\end{itemize}

Evaluations are performed primarily in the planning domain. Given an origin state and a set of goal atoms, the system must find a valid plan that fulfills all goal conditions. Natural language problem statements are used throughout to preserve generality.

\subsubsection{Naive Baseline}

The first baseline is a naive approach: prompting the LLM directly to solve the problem using only domain-specific context, instructions, the origin state, and the goal. 

Evaluated on the Blocksworld domain (five simple predicates, four actions), the primary model, Qwen-14B, struggles to find valid plans, reaching the goal in only 3 of 50 test instances. While thinking mode improves this to 23 of 50, better methods are clearly required. For reference, GPT-4o-mini completed 1 of 10 tasks correctly, and GPT-4o completed 6 of 10, likely due to its larger parameter count.
GPT-5 solves all problem instances and would need more difficult problems to evaluate.

\subsubsection{Basic Tree-of-Thought}

The second baseline is basic ToT, also evaluated on Blocksworld. ToT requires specific parameters, primarily \textit{max\_depth} and \textit{branch\_factor}. 

A primary constraint of ToT is its inherent computational inefficiency, making high-depth searches infeasible. Consequently, test instances were filtered to include only problems with solution plans of length 8 and below, allowing parameters of \textit{max\_depth} = 8 and \textit{branch\_factor} = 2. 

Beyond base parameters, ToT configurations vary across three main axes:
\begin{enumerate}
    \item Traversal method (Depth-First Search vs. Breadth-First Search).
    \item Prompt phrasing.
    \item Action/State mapping strategy: Combining action and state generation into a single abstract step (Direct) versus explicitly querying them separately (\textbf{Explicit State Action} or ESA).
\end{enumerate}

Basic ToT suffers from the same unreliability as the naive baseline. Repeated recursive prompting often triggers disastrous chain reactions, where one error degrades subsequent responses, causing the model to break output limits by repeating prompt instructions.

Without prompt extensions, the base ToT approach failed to solve a single Blocksworld instance. Implementing the targeted "pick-up vs. unstack" example and iterating on the phrasing yielded a configuration that solved 11 of 50 instances. Activating thinking mode pushed this to 94\% success, though at massive computational expense (averaging 229k tokens for normal, 330k for thinking). 

Contrary to expectations, explicitly separating the steps (\textbf{ESA}) degraded performance, taking longer to tune and ultimately performing worse than \textbf{Direct} generation, likely because increasing the volume of LLM queries per tree multiplies the opportunities for mistakes.

\subsubsection{Tree-of-Thought with Novelty Pruning}

\begin{figure}[t]
  \centering
  \includegraphics[width=1\columnwidth]{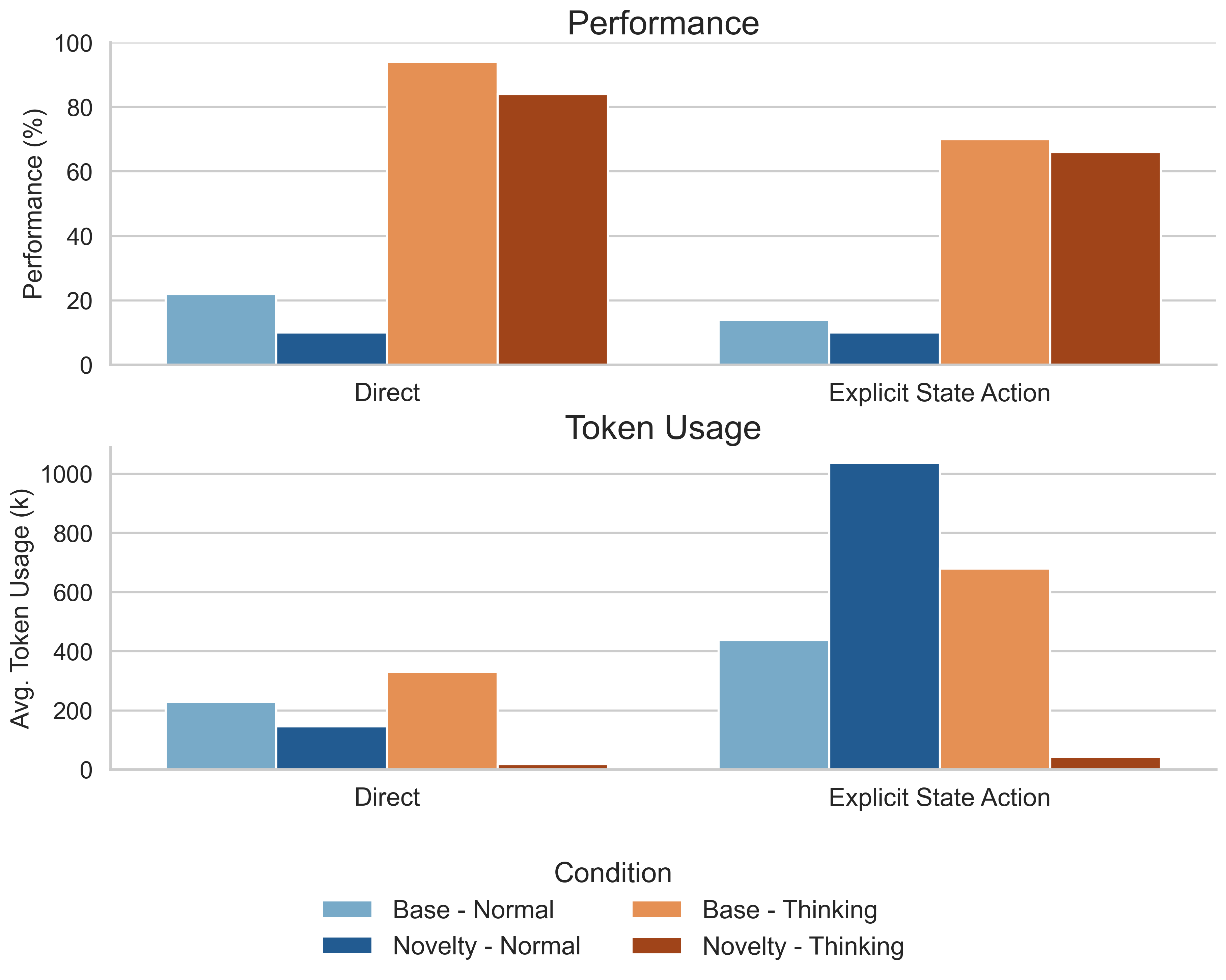}
  \caption{Comparison of performance and average token cost of Base ToT and Novelty Pruning with 50 test instances in Blocksworld (DFS).}
  \label{fig:tot_comparison}
\end{figure}

\begin{figure}[t]
  \centering
  \includegraphics[width=1\columnwidth]{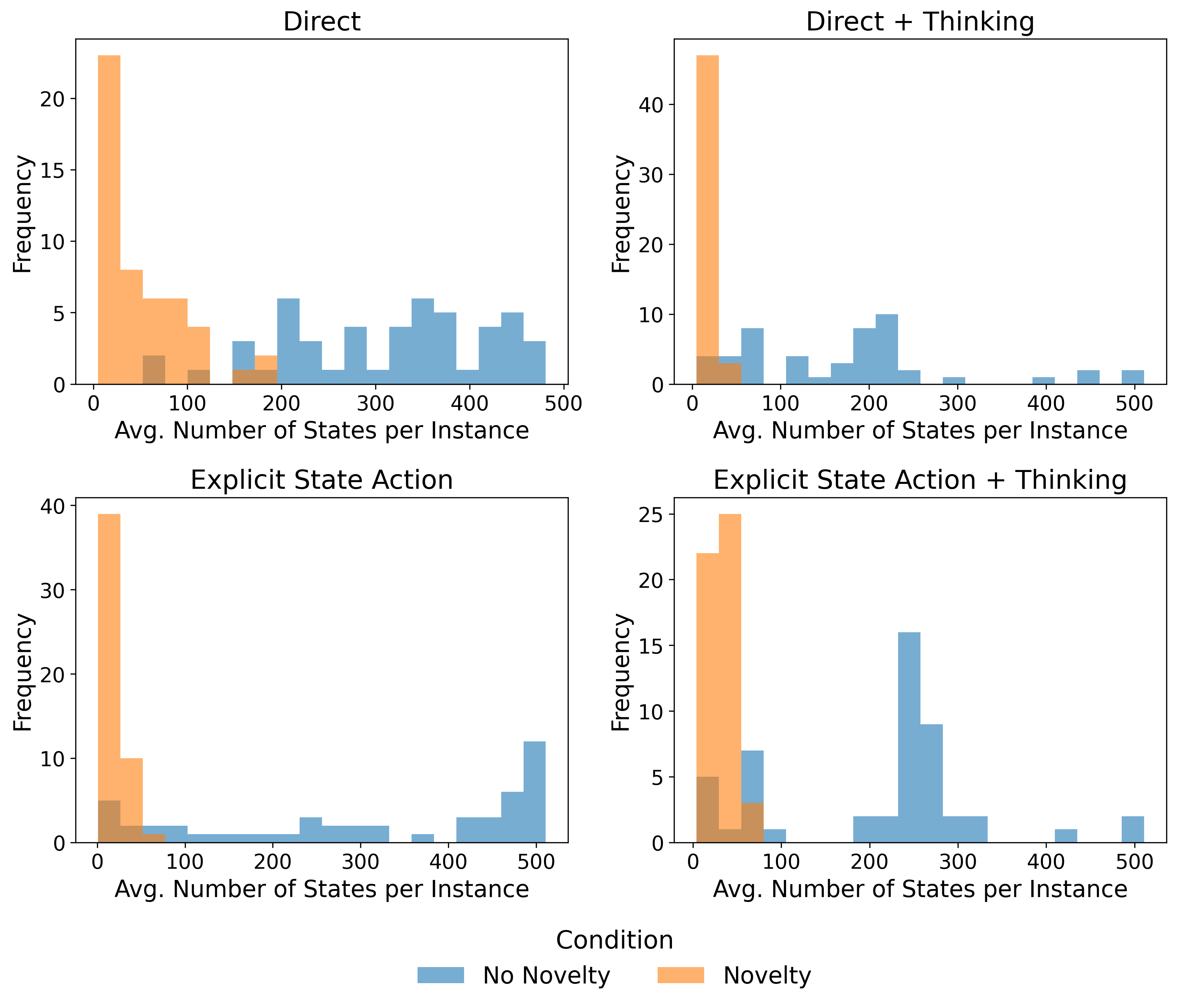}
  \caption{Distribution of average number of states per instance of different ToT configurations.}
  \label{fig:states_histograms}
\end{figure}

We now evaluate the core proposed approach: improving ToT by pruning with an LLM-driven novelty measure.
Results comparing basic ToT against Novelty Pruning are detailed in \Cref{tab:tot-comparison-blocksworld}, \Cref{fig:tot_comparison}, and \Cref{fig:states_histograms}.

The success of novelty pruning depends heavily on the underlying base performance. Configurations that performed well previously (such as Direct + Thinking) maintain their high success rates while consuming significantly fewer tokens. However, configurations with poor base performance sometimes consume \textit{more} tokens due to the added overhead of the novelty queries themselves. Across all successful runs, pruning significantly reduces the total number of generated states.

Novelty pruning also makes Breadth-First Search (BFS) computationally viable. However, as shown in \Cref{tab:tot-comparison-blocksworld}, BFS does not uniformly lower token costs compared to DFS, because BFS inherently explores and prunes a wider volume of states on average, while DFS can quickly solve easier problems without wasting compute.

\begin{table*}[!ht]
  \small
  \centering
  \begin{tabularx}{\textwidth}{l l *{8}{C}}
    \toprule
    \multirow{3}{*}{\textbf{Search}} & \multirow{3}{*}{\textbf{Method}} & \multicolumn{4}{c}{\textbf{Normal}} & \multicolumn{4}{c}{\textbf{Thinking}} \\
    \cmidrule(lr){3-6} \cmidrule(lr){7-10}
     & & \multicolumn{2}{c}{Base} & \multicolumn{2}{c}{Nov. Pruning} & \multicolumn{2}{c}{Base} & \multicolumn{2}{c}{Nov. Pruning} \\
    \cmidrule(lr){3-4} \cmidrule(lr){5-6} \cmidrule(lr){7-8} \cmidrule(lr){9-10}
     & & Perf. & ATU & Perf. & ATU & Perf. & ATU & Perf. & ATU \\
    \midrule
    \multirow{2}{*}{DFS} 
      & Direct & 7/50  & 231k  & 5/50 & 146k  & 47/50 & 330k & 42/50 & 18k \\
      & ESA    & 7/50  & 438k  & 5/50 & 1038k & 45/50 & 686k & 33/50 & 43k \\
    \addlinespace[0.5em]
    \multirow{2}{*}{BFS} 
      & Direct & 9/50  & 49k   & 4/50 & 51k   & 49/50 & 15k  & 34/50 & 29k \\
      & ESA    & 21/50 & 13k   & 5/50 & 829k  & 47/50 & 22k  & 27/50 & 89k \\
    \bottomrule
  \end{tabularx}
  \caption{Comparison of Base vs. Novelty Pruning in Blocksworld domain with DFS and BFS. ATU: Avg. Token Usage; ESA: Explicit State Action.}
  \label{tab:tot-comparison-blocksworld}
\end{table*}

All of the tests so far have been in the Blocksworld domain. To show the generalizability of the approach, the tests were run again in the also well-known Logistics domain~\cite{1998_planning_comp}. The objective in this domain is to bring one or more packages to their destination across cities using trucks and airplanes. The results for this domain can be found in \Cref{tab:tot-comparison-logistics} where we use depth-first search as it performs better for high depth domains.

\begin{table*}[!ht]
  \small
  \centering
  \begin{tabularx}{\textwidth}{l *{8}{C}}
    \toprule
    \multirow{3}{*}{\textbf{Method}} 
      & \multicolumn{4}{c}{\textbf{Normal}} 
      & \multicolumn{4}{c}{\textbf{Thinking}} \\
    \cmidrule(lr){2-5} \cmidrule(lr){6-9}
    & \multicolumn{2}{c}{Base} & \multicolumn{2}{c}{Nov. Pruning} 
    & \multicolumn{2}{c}{Base} & \multicolumn{2}{c}{Nov. Pruning} \\
    \cmidrule(lr){2-3} \cmidrule(lr){4-5} \cmidrule(lr){6-7} \cmidrule(lr){8-9}
    & Perf. & ATU & Perf. & ATU & Perf. & ATU & Perf. & ATU \\
    \midrule
    Direct & 24/50 & 101k & 19/50 & 12k & 47/50 & 69k & \textbf{48/50} & \textbf{9k} \\
    Explicit State Action & 19/50 & 100k & 24/50 & 83k & \textbf{48/50} & \textbf{204k} & 24/50 & 43k \\
    \bottomrule
  \end{tabularx}
  \caption{Comparison of Base vs. Novelty Pruning in Logistics Domain using Depth-First Search. ATU: Avg. Token Usage.}
  \label{tab:tot-comparison-logistics}
\end{table*}

As visible in the data, the approach easily generalizes to this domain with some configurations performing even better than in Blocksworld. The only manual input for the domain change was the handwritten context prompt explaining the environment. No further prompt tuning was done, showing that the prompts found previously generalize to other planning domains.

To test how well the approach transfers to domains outside of planning, all tests were run on the MATH benchmark~\cite{MATH_benchmark} using a maximum depth of 6.
This benchmark contains various math problems from high school mathematics competitions and was filtered to only contain questions of the highest difficulty (Level 5).
Due to the low depth, all tests could be run using depth-first and breadth-first search.
Again, no prompts were changed to accommodate the new problem domain.
The results can be found in \Cref{tab:tot-comparison-math}.

\begin{table*}[!ht]
  \small
  \centering
  \begin{tabularx}{\textwidth}{l l *{8}{C}}
    \toprule
    \multirow{3}{*}{\textbf{Search}} 
      & \multirow{3}{*}{\textbf{Method}} 
      & \multicolumn{4}{c}{\textbf{Normal}} 
      & \multicolumn{4}{c}{\textbf{Thinking}} \\
    \cmidrule(lr){3-6} \cmidrule(lr){7-10}
      & & \multicolumn{2}{c}{Base} & \multicolumn{2}{c}{Nov. Pruning} 
        & \multicolumn{2}{c}{Base} & \multicolumn{2}{c}{Nov. Pruning} \\
    \cmidrule(lr){3-4} \cmidrule(lr){5-6} \cmidrule(lr){7-8} \cmidrule(lr){9-10}
      & & Perf. & ATU & Perf. & ATU & Perf. & ATU & Perf. & ATU \\
    \midrule
    \multirow{2}{*}{DFS} 
      & Direct & 9/50  & 55k & 8/50  & 128k & 38/50 & 9k  & 42/50 & 10k \\
      & ESA    & 17/50 & 9k  & 8/50  & 11k  & 36/50 & 12k & 35/50 & 11k \\
    \addlinespace[0.5em]
    \multirow{2}{*}{BFS} 
      & Direct & 9/50  & 49k & 8/50  & 134k &\textbf{ 49/50} & \textbf{15k} & 45/50 & 15k \\
      & ESA    & 21/50 & 13k & 16/50 & 10k  & 47/50 & 22k & \textbf{46/50} & \textbf{21k} \\
    \bottomrule
  \end{tabularx}
  \caption{Comparison of Base vs. Novelty Pruning on MATH Benchmark (Level 5) with DFS and BFS. ATU: Avg. Token Usage; ESA: Explicit State Action.}
  \label{tab:tot-comparison-math}
\end{table*}
When thinking mode is disabled, ESA mapping slightly improves performance and reduces token costs compared to Direct prompting.
This is because the prompts for the direct method that were frozen after tuning for Blocksworld, lead to smaller reasoning steps, resulting in deeper search trees.
Boolean novelty pruning appears to have a negligible impact on both performance and token cost. While states were successfully pruned, the token savings were offset by the cost of the novelty queries themselves.
These outcomes indicate that while the method transfers out-of-the-box, achieving optimal token efficiency in distinct reasoning domains likely requires domain-specific prompt tuning.

\section{Conclusion}
\label{ch:conclusion}

The main goal of this work was to reduce the token cost of Tree of Thoughts 
(ToT) through novelty-based pruning using an LLM. The potential of this 
approach beyond classical planning was motivated through a qualitative 
evaluation on a common reasoning problem. A general implementation was 
developed that enables flexible experimentation and prompt tuning with minimal 
manual effort when switching domains.

A dedicated evaluation framework was used to examine the individual 
sub-components of the method, revealing several key challenges. The full 
approach was then benchmarked against a baseline ToT across multiple prompt 
template and parameter configurations. These configurations proved to have a 
substantial impact on both performance and cost. In well-configured settings, 
novelty pruning worked as intended: the model pruned the majority of states 
while still allowing the search to reach valid solutions, reducing token cost up to \textbf{20 times}, for same performance. In poorly configured settings, 
however, performance degraded significantly, with near-zero solve rates and 
average costs sometimes increasing by over 100\%.

This instability appears to be a fundamental characteristic of approaches 
relying on repeated or recursive LLM prompting, making configuration 
laborious and unpredictable. At the same time, such methods offer genuine 
advantages over classical planners, particularly in their adaptability to new 
domains and natural language understanding. While this work was inspired by 
width-based pruning in classical planning, the resulting method diverges 
considerably from its origin: no guarantees are made on solution quality or 
the precise computation of novelty, and the model's pruning behavior remains 
largely driven by duplicate detection rather than true novelty reasoning.

Overall, this work presents and evaluates an approach that successfully 
improves upon ToT when well-configured, while also surfacing broader 
challenges for LLM-based planning methods. Future work could investigate goal 
serialization, alternative novelty estimators, or the use of larger base 
models, which may enable deeper state understanding and more robust pruning.

\appendix

\bibliography{aaai2026}


\end{document}